\def\year{2021}\relax
\documentclass[letterpaper]{article} 
\usepackage{aaai21}  
\usepackage{times}  
\usepackage{helvet} 
\usepackage{courier}  
\usepackage[hyphens]{url}  
\usepackage{graphicx}  
\usepackage{booktabs}  
\usepackage{mathtools} 
\usepackage{natbib}  
\usepackage{caption} 
\usepackage{array}
\usepackage{caption}
\usepackage{amsthm}

\title{FreaAI: Automated extraction of data slices  to test machine learning models}
\author{Samuel Ackerman, Orna Raz, Marcel Zalmanovici}

\date{%
    IBM Research, Haifa, Israel\\[2ex]%
}
\affiliations{
    \\
    IBM Research, Haifa, Israel\\
    }

\graphicspath{ {./figures/} }

\begin{document}
\maketitle
\begin{abstract}
Machine learning (ML) solutions are prevalent. However, many challenges exist in making these solutions business-grade. One major challenge is to ensure that the ML solution provides its expected business value. In order to do that, one has to bridge the gap between the way ML model performance is measured and the solution requirements.
In previous work \cite{barash2019bridging} we demonstrated the effectiveness of utilizing feature models in bridging this gap. Whereas ML performance metrics, such as the accuracy or F1-score of a classifier, typically measure the average ML performance, feature models shed light on explainable data slices that are too far from that average, and therefore might indicate unsatisfied requirements. For example, the overall accuracy of a bank text terms classifier may be very high, say $98\% \pm 2\%$, yet it might perform poorly for terms that include short descriptions and originate from commercial accounts. A business requirement, which may be implicit in the training data, may be to perform well regardless of the type of account and length of the description. Therefore, the under-performing data slice that includes short descriptions and commercial accounts suggests poorly-met requirements. In this paper we show the feasibility of automatically extracting feature models that result in explainable data slices over which the ML solution under-performs. 
Our novel technique, IBM FreaAI aka FreaAI, extracts such slices from structured ML test data or any other labeled data. We demonstrate that FreaAI can automatically produce explainable and statistically-significant data slices over seven open datasets. 
\end{abstract}
\section{Introduction}\label{sec:intro}
Software systems that utilize ML are becoming more and more prevalent. The quality of ML-based solutions greatly depends on the data used to train a solution. Common approaches for assessing the quality of trained ML solutions (we will refer to to these as ML) use the average performance of the ML according to some metric, such as accuracy or F1-score for a classifier and $R^2$ or root mean square error (RMSE) for regression, over unseen yet representative data, such as that of a test set. Of course, even if these average measures are satisfactory, there are likely to exist data records for which the ML performance is far below the average. It is important to understand this behavior as these far-away records might be indicative of unmet requirements. For example, the third row in Table~\ref{tab:slice_ex} indicates that the ML model trained over loan approval data (the \textbf{Adult} dataset) under-performs for people in the ages of 33 to 64. While the overall accuracy is roughly 85\%, the accuracy for people in this age group is roughly 80\%. Probably this fails to meet a requirement of providing similar ML performance regardless of age. Another example in that table is under-performance of roughly 67\% accuracy for people who work 40 to 43 hours per week and belong to workclass category 5. It is probably required that the solution be insensitive to the number of hours a person works and their workclass. These are examples of `data slices'. The first is a single feature data slice and the second is a 2-way feature interaction data slice.  

In previous work \cite{barash2019bridging} we showed that data slices, defined by a Combinatorial Testing (CT) feature model, result in  identifying far-away records in a way that a user can understand and act upon. In that work the feature models were defined manually, and included various abstractions of the ML input data as well as additional metadata. In this work we concentrate on automatically creating feature models for one simple yet useful type of abstraction---that of binning input data feature values. The next section provides background about feature models.

While searching over single categorical feature values to find interesting slices is straightforward, searching over continuous-valued features and over interactions of multiple features is challenging. An exhaustive search is infeasible as the search space is exponential. Moreover, reporting such records or data slices should be done such that the user can take corrective actions. This means that the user needs to understand what defines a data slice. FreaAI is our novel technology that implements a set of heuristics to efficiently suggest data slices that are, by construction, explainable, correct, and statistically significant.
`Explainable' means that data slices are described solely as ranges or single values and their unions. 
FreaAI is named after the Norse goddess Frea (spelling varies) who, among other traits, can foresee the future, but does not share this knowledge with humans. Our technology suggests data slices that are correct, yet the results may be incomplete in that not all possible under-performing slices may be found. 

The contribution of this paper is a set of heuristics for automatically defining feature models, such that each slice in that model is indicative of ML under-performance, is statistically significant, and can be easily understood by a human.
We validate these heuristics by analyzing seven open datasets. FreaAI generates data slices that contain significantly more mis-predicted data records than would be expected by randomly drawing the same number of data records. 

We next list related work and provide background from our previous work about feature models and data slices. The methodology of heuristically and automatically suggesting data slices and the definition of our requirements from data slices are then presented. The experimental results of running our technology on seven open data-sets follow, and we conclude with a discussion and brief summary of the contributions of this work.

\section{Related work and background} \label{sec:related}
Several works capture challenges and best practices in developing ML solutions, e.g., \cite{Google_technical_debt,Google_testscore,Google_43rules}, or suggest organizational work-flows for building successful AI projects \cite{NgAITransformation}.

We summarize the most relevant background and results 
from our previous work~\cite{barash2019bridging}. The work we report on in here is motivated by that work.
In a nutshell, we suggested the use of CT modeling methodology for validating ML solutions. Building on that work, here we report on FreaAI, a novel technology for automatically extracting feature models, which are an instance of CT models, out of ML data, implementing an abstraction of automated binning according to ML performance. 

Following are more details on our previous work. 
We demonstrated that the
Combinatorial modeling methodology is a viable approach for gaining actionable insights on the quality of ML solutions. Applying the approach results in identifying coverage gaps indicating uncovered requirements. The approach also identifies under-performing slices, suggesting areas where the current ML model cannot be relied upon. When used in the development stage of the ML solution life-cycle, the approach can also suggest additional training data or the inclusion of multiple ML model variants as part of the overall ML solution.  
The methodology may also assist in directing the labeling effort to areas that would best cover the business requirements. 
That work utilized the modeling process used by CT ~\cite{Burroughs94,Cohen:2006:TAC,Dalal:1999,Grindal:2006:ECS,Kuhn:2013:ICT,Williams:2000,CHARM2014} to detect weak business areas of the ML solution and strengthen them.

CT is a well-known testing technique to achieve a small and effective test plan.
The systematic combinatorial modeling methodology used by CT provides insight on the business requirements and ties them to testing artifacts. 
We utilized the combinatorial model as a means to detect gaps and weaknesses of the ML solution. Business requirements often do not appear as features in the ML solution training data. Rather, they can be defined as abstractions over these features or may utilize additional information that is not always available for learning.  The CT model parameters may then include abstractions as simple as dividing values into bins, such as defining ranges over age as young, middle-aged and elderly, as well as more complex abstractions such as defining mathematical functions over the relation between multiple features. 

Using the mapping between CT model parameters and ML model features, we defined {\it data slices} over the training data as follows. We observed that a $t$-way interaction coverage requirement induces a division of the training data into slices. Each slice corresponds to a specific value combination of size $t$ and consists of all data records that map to that specific value combination. Naturally, the number of data slices is equal to the number of $t$-way value combinations in the CT model. Note that the slices are not disjoint, since each data record maps to multiple value combinations of size $t$, reflecting different business requirements that it captures. Data slices over credit data may include records that belong to young women (a 1-way interaction slice) and records that belong to young women with only high-school level education (a 2-way interaction slice).
We performed coverage gap analysis of the training data associated with each data slice. Slices with no corresponding training data expose business areas in which the ML solution will likely under-perform. For slices with sufficient data or support, we computed the ML metric over each slice. We expected the result for a slice to be within the confidence interval of the ML metric over the entire data. If that was not the case, we highlighted the slice as indicative of business requirements that are not well addressed by the ML solution. 

The work we report on here provides a set of heuristics, FreaAI, for applying the above CT modeling methodology on the ML solution test data in order to better validate it and verify that it addresses its requirements.
As even the simple abstractions were shown to be useful, FreaAI was designed to automatically generate simple abstractions such as binning of continuous features and identification of categorical feature values and combinations of feature. The resulting data slices are guaranteed to: (1) under-perform relative to the overall ML solution performance, (2) be explainable, (3) be statistically significant. Details on the heuristics and how they guarantee these properties follow in the Methodology Section. 
\section{Methodology}\label{sec:methodology}

FreaAI automatically analyzes features and feature interactions in order to define feature models in which each of the data slices under-performs.
It requires only the test data and the ML prediction for each record, without requiring knowledge of the ML model, for instance.

\subsection{Single feature analysis}
FreaAI starts by analyzing single features. The rationale is that the simpler the reason for an issue, the easier it is to understand its cause and, as a result, easier to come up with a fix.
FreaAI heuristically decides which features should be treated as categorical and which as continuous and does so automatically based on the type of feature data and number of unique values. Alternatively, the user can supply this information. Looking for under-performing data slices over single categorical features is simple. FreaAI computes the ML performance of data slices that are defined by each of the feature possible values (categories). Finding data slices over single continuous features is challenging, as we need a way to find interesting ranges. We do that in various ways, explained in the following paragraphs.

\subsection{Feature interactions analysis}
 FreaAI also analyzes feature interactions. It does so in two ways: (1) by slicing on a category or range found when analyzing a single feature and repeating the process for a single feature on that subset, or (2) by using a method which can divide the search space on multiple axis at once---in our case, a decision tree (DT) on \textit{n} features. From CT we know that most software problems involve a small number of variables. Empirically, we notice that this seems to be the case for data features used for ML as well; single features and interactions of two or three features already uncover many problems. An inherent problem with increasing the interaction level is that the higher the interaction level, the less likely it is to have sufficient support.
 
 \subsection{Heuristics for feature and interaction analysis}
 We experimented with multiple feature-analysis techniques.
    These included clustering methods, building decision trees, and Highest Posterior Density (HPD), also known as High Density Regions (HDR)~\cite{high_density_regions} to extract feature models with under-performing slices. We shortly describe each of the techniques. We found decision trees and HPD to be the most effective, therefore we report only on the results of applying these heuristics in the Experimental Results section. 
 
\subsubsection{Clustering as a feature analysis heuristic}
 We applied clustering techniques to group the test data. We utilized multiple clustering algorithms that are inherently different in concept, such as k-means and DBScan. We then tested whether any of the resulting clusters 
 exhibited ML under-performance. We also varied the clustering metric and the parameters to get different clusters. While this method worked well for identifying under-performing slices, it was hard to automatically point out the offending feature(s) in a way that was user-actionable. We could only mark a (small) subset of features as being more dominant in the under-performing cluster than in the rest of the test set, suggesting one or more of them could be responsible. 
 
 \subsubsection{HPD as a feature analysis heuristic}
\begin{figure}
  \includegraphics[width=\linewidth]{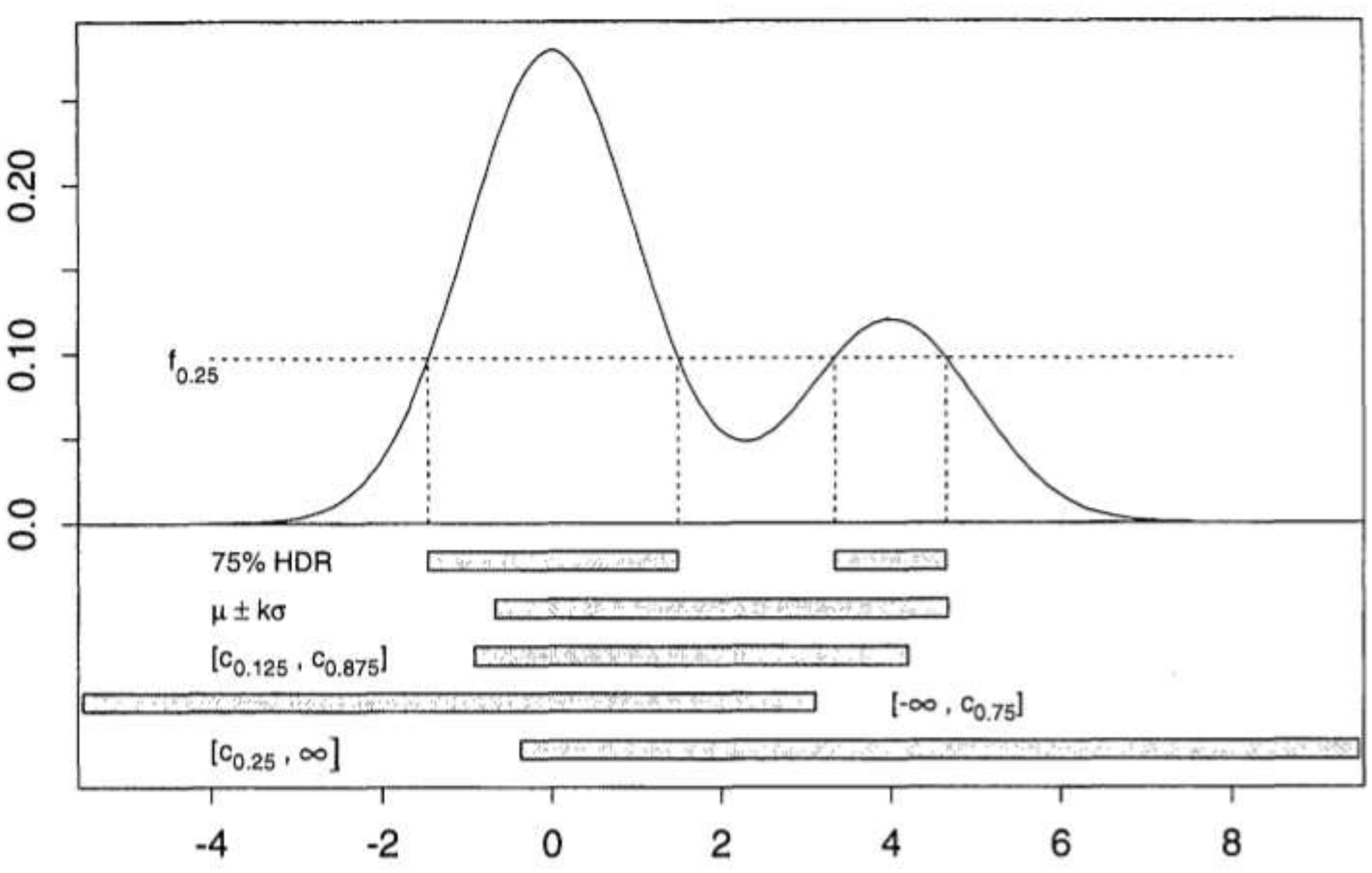}
  \caption{Illustration of five different intervals containing 75\% of the density from an example (bimodal) empirical sample (source: ~\cite{high_density_regions}).  The first one is the HDR (or HPD) since of the five, it is the shortest (highest-density); in this case, the HPD is given by a union of two non-overlapping intervals due to the bimodality of the distribution.}
  \label{fig:HDR_figure}
\end{figure}

 HPD is a nonparametric method that computes the shortest possible interval or union of non-overlapping intervals containing a given proportion of records in a univariate and numeric sample; it essentially gives the `best possible' empirical confidence interval for a sample.  An illustration from the original paper is given in Figure~\ref{fig:HDR_figure}.   
 
 Since HPDs can only be used on univariate numeric data (unlike decision trees, which do not have such restrictions), we use it in our heuristic only to identify interesting slices for each numeric feature alone (i.e., no higher-order interactions).  So the slices are explainable and to avoid potential overfitting due to multi-modality of the distribution, we furthermore restrict the HPD to consist of a single interval, rather than a union.  For the sample of each relevant feature in the test dataset, we begin by finding the initial HPD, and iteratively shrink that area by some epsilon (typically 0.05) checking whether the performance on the new sample increases or decreases relative to the previous one. If it decreases, then we have found a smaller range which under-performs. Alternatively, if it increases, then at least one of the discarded ranges (one on either side) under-performs. This process is repeated after dropping the highest-density range from the feature until we remain with too little data, e.g. 10\% of the data records.
 
\begin{figure}[ht]
    \caption{Shows the areas discarded when changing HPD to include 85\% [\mbox{-0.04553} to 0.26878 with a performance of 0.9858] instead of 90\% {[\mbox{-0.07738} to 0.27257 with 0.9840]} density. The two marked bars are the areas FreaAI checks when the performance of the lower density is higher. This data belongs to feature `\textit{MFCCs\_22}' from \textbf{Anuran (family)}.}
    \label{fig:hpd_example}
    \centering
    \includegraphics[width=0.45\textwidth]{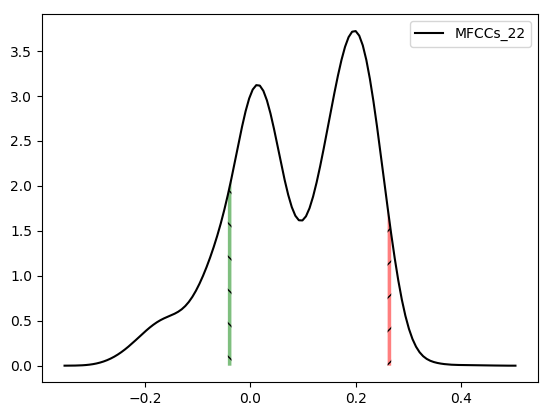}
\end{figure}

 Figure~\ref{fig:hpd_example} depicts one iteration of the HPD heuristic over one feature, `\textit{MFCCs\_22}' from the \textbf{Anuran (family)} dataset. First, the area under the probability density curve, up to and including the vertical ranges, is identified by HPD as containing 90\% of the probability density. Our HPD heuristic iteration shrinks that area by 5\% by removing the vertical ranges and re-checks the performance of the updated range. It then proceeds as described above. 
 
 \subsubsection{Decision Trees as a feature analysis heuristic}
 The decision trees heuristic works as follows. For each subset of one, two or three features the heuristic defines a decision tree classification problem using these features as input. The output is always defined by a binary target of True/False. `True' means that the original ML model predicted the original output successfully and `False' means that it failed in its prediction. Notice that this makes each classification problem binary, regardless of the number of original labels. The decision tree heuristic then fits a decision tree model to each of the above training data, but ignores the performance of the resulting model. Instead, it walks the generated tree looking for splits where the purity of `False' is high enough for the slice defined by it to be under-performing and keeps only those. 
 
 \subsubsection{Minimal support}
 The slicing heuristics consider as candidates only slices that have at least minimum support. 
Minimum support is heuristically calculated as the maximum between 2 and 0.5\% of mis-predicted instances in the test dataset. The rationale is that in a dataset with a high metric performance there will be few mis-predictions, therefore a slice with a small number of mis-predictions is important, whereas for a model whose performance is relatively low there will be many mis-predictions requiring a slice to have more mis-predictions to be interesting. In addition, by taking a percentage we automatically adjust to different test dataset sizes, e.g., Table~\ref{tab:data_set_info} shows that the \textbf{Statlog} dataset has 300 records, while the \textbf{Adult} dataset has over 14,000 records. The percentage is user-configurable. However, in our experiments the empirical values that we chose were satisfactory. 

 \subsection{Defining data slices requirements}
 We define the following requirements from the data slices reported by FreaAI, regardless of the heuristic that generated a data slice. A data slice must be 
 \begin{enumerate}
     \item correct, 
     \item statistically significant, and 
     \item explainable.
 \end{enumerate}
  We use a methodology that abides to these requirements and generates slices which are correct-by-construction. The slices are correct because they are subsets of the test set for which we run the same performance metric as on the original dataset. They are statistically significant because we take only slices where the performance is considerably (4\% by default) lower than the expected lower-bound of the confidence interval of the performance metric. FreaAI further reports only on data slices that have both minimal support and a sufficiently low p-value, as computed by the hypergeometric distribution test, described in the Experimental Results section. The slices are also self-explainable since they are generated from an easily defined range on a small number of features. Moreover, since we focus on structured datasets with engineered features, the ML model developer should be able to draw useful conclusions. Table~\ref{tab:slice_ex} lists a few examples of data slices. These are explainable by the way they are defined. For example, the first data slice indicates a problem in the Adult ML model performance when the \textit{`relationship'} category is $10$. This is an example of a 1-way or single feature data slice. Another data slice indicates a problem when the feature \textit{`hours\_per\_week'} has values between $40$ and $43$ and the feature \textit{`workclass'} category is $5$. This is an example of a 2-way or feature interactions data slice.

\section{Experimental results}
\label{sec:eval}

Our experiments demonstrate the ability of our heuristics, as implemented by FreaAI, to find under-performing data slices that are correct, statistically significant, and explainable. 

\subsection{Input}\label{subsec:exp_input}
We tested FreaAI on open-source data sets, six taken from UCI and one from ProPublica. The datasets are \textbf{Adult}, \textbf{Avila}, \textbf{Anuran} (Family, Genus, Species targets), \textbf{statlog} (German credit data) from UCI and \textbf{COMPAS Recidivism Risk Score} from ProPublica. 

Table \ref{tab:data_set_info} lists some general information about the datasets and the performance of the models trained on them. We made no attempt at creating the best possible model for each dataset as it is not the goal of this work. We trained a random forest model for each dataset and got ML performance that was similar to the reported performance of each in the literature. 






\begin{table}[htb]
  \centering
  \caption{Information about the datasets that we tested, listing the number of \textbf{test} records, the model performance measured by accuracy and the low/high bounds of the confidence interval for the accuracy}
    \begin{tabular}{l@{\hspace{.7\tabcolsep}}r@{\hspace{.7\tabcolsep}}r@{\hspace{.8\tabcolsep}}r@{\hspace{.7\tabcolsep}}r@{\hspace{.7\tabcolsep}}}
    \toprule
    \textbf{Dataset} & \textbf{\# records} & \textbf{Acc} & \textbf{Low CI} & \textbf{High CI} \\
    \midrule
    \textbf{Adult}              & 14653 & 0.852 & 0.845 & 0.857 \\
    \textbf{Avila}              & 6169  & 0.979 & 0.976 & 0.983 \\
    \textbf{Anuran (family)}    & 2159  & 0.980  & 0.974 & 0.987 \\
    \textbf{Anuran (genus)}      & 2159  & 0.967 & 0.959 & 0.974 \\
    \textbf{Anuran (species)}   & 2159  & 0.969 & 0.962 & 0.977 \\
    \textbf{Statlog}            &  300  & 0.763 & 0.708 & 0.821 \\
    \textbf{ProPublica}         & 1852  & 0.742 & 0.718 & 0.759 \\
    \bottomrule
    \end{tabular}%
  \label{tab:data_set_info}%
\end{table}%

In our experiments, FreaAI gets as input a test dataset. This is data that was not used for training the ML model, is considered representative of the underlying data distribution, and contains both the ground truth value of the target feature as well as the ML model output, either a numeric or class prediction as appropriate.

\subsection{Finding under-performing data slices}\label{subsec:exp_slices}
FreaAI implements various slicing heuristics to find candidate under-performing data slices, as discussed in the Methodology Section. FreaAI computes the performance by comparing the ground truth and the ML model prediction that are provided as part of the input data, according to the same ML performance metric used to assess the ML model performance. For simplicity, in all the experiments that we report on here we use accuracy as the ML performance metric and treat all features as numeric. As the Methodology Section explains, we report on the data slices resulting from applying the HPD and the DT slicing heuristics. 

FreaAI's heuristics consider as candidates only slices that have at least minimum support, as the Methodology section describes. 
The candidate slices are further filtered according to their statistical significance and FreaAI reports only on statistically significant slices. The next subsection provides details about the hypergeometric distribution test that FreaAI applies to achieve that. 

Tables~\ref{tab:hpd_support_1way}--\ref{tab:dt_support_2way} show that regardless of the heuristic used to calculate the slices (HPD or DT) or the level of feature interaction (1-way or 2-way), the support varies greatly for all datasets, with the occasional exception of the \textbf{Statlog} dataset which is the smallest dataset (see dataset statistics in Table~\ref{tab:data_set_info}). This is to be expected, as our goal is to find groups of mis-predicted records for which we can automatically provide a readable explanation. Table~\ref{tab:slice_ex} provides examples from the Adult dataset of under-performing data slices. 

It is highly encouraging that our heuristics consistently find under-performing data slices. Moreover, the reported slices often have substantial support, increasing the usefulness of the report to the user. 

\begin{table}[htb]
  \centering
  \caption{Summary of the number of under-performing slices for 1 and 2-way feature interaction and their total. `cand' indicates the candidates found the \textbf{HPD} heuristic. `rep' indicates how many candidate slices were actually reported out of the `cand' list after applying the hypergeometric distribution test with p-value $<$ 0.05.}
    \begin{tabular}{l@{\hspace{.7\tabcolsep}}r@{\hspace{.7\tabcolsep}}r@{\hspace{.7\tabcolsep}}r@{\hspace{.7\tabcolsep}}r@{\hspace{.7\tabcolsep}}r@{\hspace{.7\tabcolsep}}r@{\hspace{.7\tabcolsep}}}
    \toprule
    \textbf{} & \multicolumn{2}{c}{\textbf{\# 1-way}} & \multicolumn{2}{c}{\textbf{\# 2-way}} & \multicolumn{2}{c}{\textbf{\# total}} \\
    \textbf{Dataset} & \textbf{cand} & \textbf{rep} & \textbf{cand} & \textbf{rep} & \textbf{cand} & \textbf{rep} \\
    \midrule
    \textbf{Adult}              & 26    & 25    & 82    & 82    & \textbf{108}   & \textbf{107} \\
    \textbf{Avila}              & 38    & 29    & 17    & 17    & \textbf{55}    & \textbf{46} \\
    \textbf{Anuran (family)}    & 80    & 58    & 386   & 371   & \textbf{466}   & \textbf{429} \\
    \textbf{Anuran (genus)}     & 105   & 61    & 723   & 638   & \textbf{828}   & \textbf{699} \\
    \textbf{Anuran (species)}   & 91    & 64    & 653   & 569   & \textbf{744}   & \textbf{633} \\
    \textbf{Statlog}            & 24    & 7     & 56    & 25    & \textbf{80}    & \textbf{32} \\
    \textbf{ProPublica}         & 12    & 4     & 13    & 7     & \textbf{25}    & \textbf{11} \\
    \bottomrule
    \end{tabular}%
  \label{tab:hpd_report}%
\end{table}%

\begin{table}[htb]
  \centering
  \caption{Summary of the number of under-performing slices for 1 and 2-way feature interaction and their total. `cand' indicates the candidates found the \textbf{DT} (decision trees) heuristic. `rep' indicates how many candidate slices were actually reported out of the `cand' list after applying the hypergeometric distribution test with p-value $<$ 0.05.}
    \begin{tabular}{l@{\hspace{.7\tabcolsep}}r@{\hspace{.7\tabcolsep}}r@{\hspace{.7\tabcolsep}}r@{\hspace{.7\tabcolsep}}r@{\hspace{.7\tabcolsep}}r@{\hspace{.7\tabcolsep}}r@{\hspace{.7\tabcolsep}}}
    \toprule
    \textbf{} & \multicolumn{2}{c}{\textbf{\# 1-way}} & \multicolumn{2}{c}{\textbf{\# 2-way}} & \multicolumn{2}{c}{\textbf{\# total}} \\
    \textbf{Dataset} & \textbf{cand} & \textbf{rep} & \textbf{cand} & \textbf{rep} & \textbf{cand} & \textbf{rep} \\
    \midrule
    \textbf{Adult}              & 36    & 35    & 285   & 271   & \textbf{321}   & \textbf{306} \\
    \textbf{Avila}              & 120   & 84    & 928   & 689   & \textbf{1048}  & \textbf{773} \\
    \textbf{Anuran (family)}    & 148   & 113   & 1539  & 1265  & \textbf{1687}  & \textbf{1378} \\
    \textbf{Anuran (genus)}     & 209   & 142   & 2123  & 1498  & \textbf{2332}  & \textbf{1640} \\
    \textbf{Anuran (species)}   & 205   & 134   & 1881  & 1336  & \textbf{2086}  & \textbf{1470} \\
    \textbf{Statlog}            & 29    & 8     & 610   & 177   & \textbf{639}   & \textbf{185} \\
    \textbf{ProPublica}         & 1     & 1     & 24    & 22    & \textbf{25}    & \textbf{23} \\
    \bottomrule
    \end{tabular}%
  \label{tab:dt_report}%
\end{table}%

\begin{table}[htb]
  \centering
  \caption{Summary of the support for 1-way HPD slices using a stricter p-value in the hypergeometric distribution test of 0.01 as threshold. The first column lists the number of reported slices. Then for each dataset the minimal (`MIN'), average (`AVG'), maximal (`MAX') and standard deviation (`STD') for the reported slices.}
    \begin{tabular}{l@{\hspace{.7\tabcolsep}}r@{\hspace{.7\tabcolsep}}r@{\hspace{.7\tabcolsep}}r@{\hspace{.7\tabcolsep}}r@{\hspace{.7\tabcolsep}}r@{\hspace{.7\tabcolsep}}}
    \toprule
    \textbf{Dataset} & \textbf{\#} & \textbf{MIN} & \textbf{AVG} & \textbf{MAX} & \textbf{STD} \\ 
    \midrule
    \textbf{Adult}              & 19 & 108 & 2601.5 & 9888  & 2931   \\
    \textbf{Avila}              & 13 & 6   & 121.1  & 305   & 77.6   \\
    \textbf{Anuran (family)}    & 32 & 2   & 109.3  & 197   & 51.9   \\
    \textbf{Anuran (genus)}     & 36 & 3   & 161.1  & 351   & 70.2   \\
    \textbf{Anuran (species)}   & 30 & 3   & 201.6  & 460   & 111.3  \\
    \textbf{Statlog}            & 2  & 3   & 78.5   & 80    & 1.5    \\
    \textbf{ProPublica}         & 3  & 24  & 485.7  & 942   & 341.9  \\
    \bottomrule
    \end{tabular}%
  \label{tab:hpd_support_1way}%
\end{table}%

\begin{table}[htb]
  \centering
  \caption{Summary of the support for 2-way HPD slices using a stricter p-value of 0.01 as a threshold.}
    \begin{tabular}{l@{\hspace{.7\tabcolsep}}r@{\hspace{.7\tabcolsep}}r@{\hspace{.7\tabcolsep}}r@{\hspace{.7\tabcolsep}}r@{\hspace{.7\tabcolsep}}r@{\hspace{.7\tabcolsep}}}
    \toprule
    \textbf{Dataset} & \textbf{\#} & \textbf{MIN} & \textbf{AVG} & \textbf{MAX} & \textbf{STD} \\ 
    \midrule
    \textbf{Adult}              & 81    & 108 & 1278.7 & 8749  & 1741.6 \\
    \textbf{Avila}              & 10    & 6   & 28.2  & 137   & 37 \\
    \textbf{Anuran (family)}    & 280   & 2   & 12.92 & 92    & 13.8 \\
    \textbf{Anuran (genus)}     & 351   & 3   & 23    & 114   & 22.7 \\
    \textbf{Anuran (species)}   & 391   & 3   & 31.1  & 231   & 37.5 \\
    \textbf{Statlog}            & 11    & 3   & 11.82 & 29    & 7.1 \\
    \textbf{ProPublica}         & 3     & 24  & 143   & 285   & 102.8 \\
    \bottomrule
    \end{tabular}%
  \label{tab:hpd_support_2way}%
\end{table}%

\begin{table}[htb]
  \centering
  \caption{Summary of the support for 1-way DT slices using a stricter p-value of 0.01 as a threshold.}
    \begin{tabular}{l@{\hspace{.7\tabcolsep}}r@{\hspace{.7\tabcolsep}}r@{\hspace{.7\tabcolsep}}r@{\hspace{.7\tabcolsep}}r@{\hspace{.7\tabcolsep}}r@{\hspace{.7\tabcolsep}}}
    \toprule
    \textbf{Dataset} & \textbf{\#} & \textbf{MIN} & \textbf{AVG} & \textbf{MAX} & \textbf{STD} \\ 
    \midrule
    \textbf{Adult}              & 22    & 108   & 2301.5 & 9941  & 2869 \\
    \textbf{Avila}              & 44    & 6     & 55.8  & 236   & 51.4 \\
    \textbf{Anuran (family)}    & 70    & 2     & 54.6  & 262   & 57.9 \\
    \textbf{Anuran (genus)}     & 72    & 3     & 74    & 348   & 78.4 \\
    \textbf{Anuran (species)}   & 71    & 3     & 73.4  & 451   & 100.1 \\
    \textbf{Statlog}            & 4     & 3     & 80    & 124   & 30.1 \\
    \textbf{ProPublica}         & 1     & 24    & 519   & 519   & 0 \\
    \bottomrule
    \end{tabular}%
  \label{tab:dt_support_1way}%
\end{table}%

\begin{table}[htb]
  \centering
  \caption{Summary of the support for 2-way DT slices using a stricter p-value of 0.01 as a threshold.}
    \begin{tabular}{l@{\hspace{.7\tabcolsep}}r@{\hspace{.7\tabcolsep}}r@{\hspace{.7\tabcolsep}}r@{\hspace{.7\tabcolsep}}r@{\hspace{.7\tabcolsep}}r@{\hspace{.7\tabcolsep}}}
    \toprule
    \textbf{Dataset} & \textbf{\#} & \textbf{MIN} & \textbf{AVG} & \textbf{MAX} & \textbf{STD} \\ 
    \midrule
    \textbf{Adult}              & 202   & 108   & 2682.5 & 10736 & 3086.8 \\
    \textbf{Avila}              & 368   & 6     & 75.4  & 366   & 66.9 \\
    \textbf{Anuran (family)}    & 870   & 2     & 71    & 366   & 70.3 \\
    \textbf{Anuran (genus)}     & 934   & 3     & 92.3  & 555   & 90.3 \\
    \textbf{Anuran (species)}   & 818   & 3     & 104.1 & 582   & 111 \\
    \textbf{Statlog}            & 86    & 3     & 87    & 176   & 41.2 \\
    \textbf{ProPublica}         & 17    & 24    & 372.9 & 826   & 220.4 \\
    \bottomrule
    \end{tabular}%
  \label{tab:dt_support_2way}%
\end{table}%

\begin{table}[htb]
  \centering
  \caption{Examples of under-performing slices taken from both the HPD and the DT heuristics on the \textbf{Adult} dataset. The average performance for \textbf{Adult} is roughly 0.85. The first 4 rows are examples of 1-way under-performing slices and the last 3 of 2-way slices. The first two rows represent categorical features, the next two continuous features, as decided by heuristic, and the following three come from the union of continuous feature ranges. Note that the last row represents a subset of the range in the third row, i.e. a slice on a secondary feature.}
    \begin{tabular}{l@{\hspace{.9\tabcolsep}}l@{\hspace{.7\tabcolsep}}r@{\hspace{.7\tabcolsep}}r@{\hspace{.7\tabcolsep}}r@{\hspace{.7\tabcolsep}}}
    \toprule
    \textbf{ATTR} & \textbf{VALUE} & \textbf{SUP} & \textbf{PERF} & \textbf{p-val} \\
    \midrule
    relationship    & 5             & 692   & 0.699 & 1.2E-25 \\
    education       & 10            & 174   & 0.707 & 7.4E-07 \\
    age             & 33-64         & 8538  & 0.797 & 1E-113 \\
    capital\_gain   & 3942-4934     & 122   & 0.688 & 3.8E-06 \\
    (hours\_per\_week,  & 40-43,        & 141   & 0.673 & 8.2E-08 \\
    \multicolumn{1}{r}{workclass)}      & \multicolumn{1}{r}{5}   &  &  &  \\
    (education\_num,    & (15, 16),     & 122   & 0.680 & 1.4E-06 \\
    \multicolumn{1}{r}{relationship)}   & \multicolumn{1}{r}{(1, 2, 3, 4)}  &  &  &  \\
    (age,                 & 33-64,       & 3019   & 0.749 & 1.3E-62 \\
    \multicolumn{1}{r}{hours\_per\_week)}         & \multicolumn{1}{r}{41-99}  & &  & \\
    \bottomrule
    \end{tabular}%
  \label{tab:slice_ex}%
\end{table}%


\subsection{Slice significance testing}
Our combinatorial procedure, along with the HPD and DT heuristics, return data slices of combinations of 1--3 features, that is subsets of the test set records, for which the model performance is worse than the performance on the test set overall.  The performance on these slices may not be significantly worse, however.  That is, in the case of a classifier, the proportion of correctly-classified observations in these subsets may not be statistically significantly lower than the proportion in the test set overall.  Due to the number of potential combinations, which grows polynomially with the number of features, we wish to only return to the user slices with statistically-significantly worse performance than the ML model overall performance.

To assess this, let the test dataset consist of $N$ records, of which $K\in\{0,\dots,N\}$ are correctly classified.  Consider a slice of $n\in \{1,\dots,N\}$ records, of which $k \in \{0,\dots,n\}$ are correctly classified.  The model under-performance (accuracy $k/n$) on the slice is considered significant if $k$ is low relative to $n\frac{K}{N}$, the expected number of correctly-classified observation in a randomly drawn subset the same size $n$ from the test set, which has overall accuracy $K/N$.  

\subsubsection{The hypergeometric distribution test}
We use the discrete statistical hypergeometric distribution, which in a general setup models the probability of a certain number of binary outcomes ($k$) from a random draw without replacement of a certain size ($n$) from a finite population (of size $N$).  In our work, we are concerned with how many correct classifications (a binary outcome) are expected in a random observation subset of a given size ($n$) from the test set (size $N$) with a total number $K$ of correct classifications.  This likelihood is the probability mass function of the hypergeometric distribution with parameters $N,\:n,\:K$ which are observed in the slice, that is, $\text{Pr}(X=x \mid N,n,K)$.  The lower-tailed p-value, which evaluates the significance of our result, is simply $\sum_{x=0}^k \text{Pr}(X=x \mid N,n,K)$, below the observed correct classification count $k$ in the slice.

\begin{figure}
  \includegraphics[width=\linewidth]{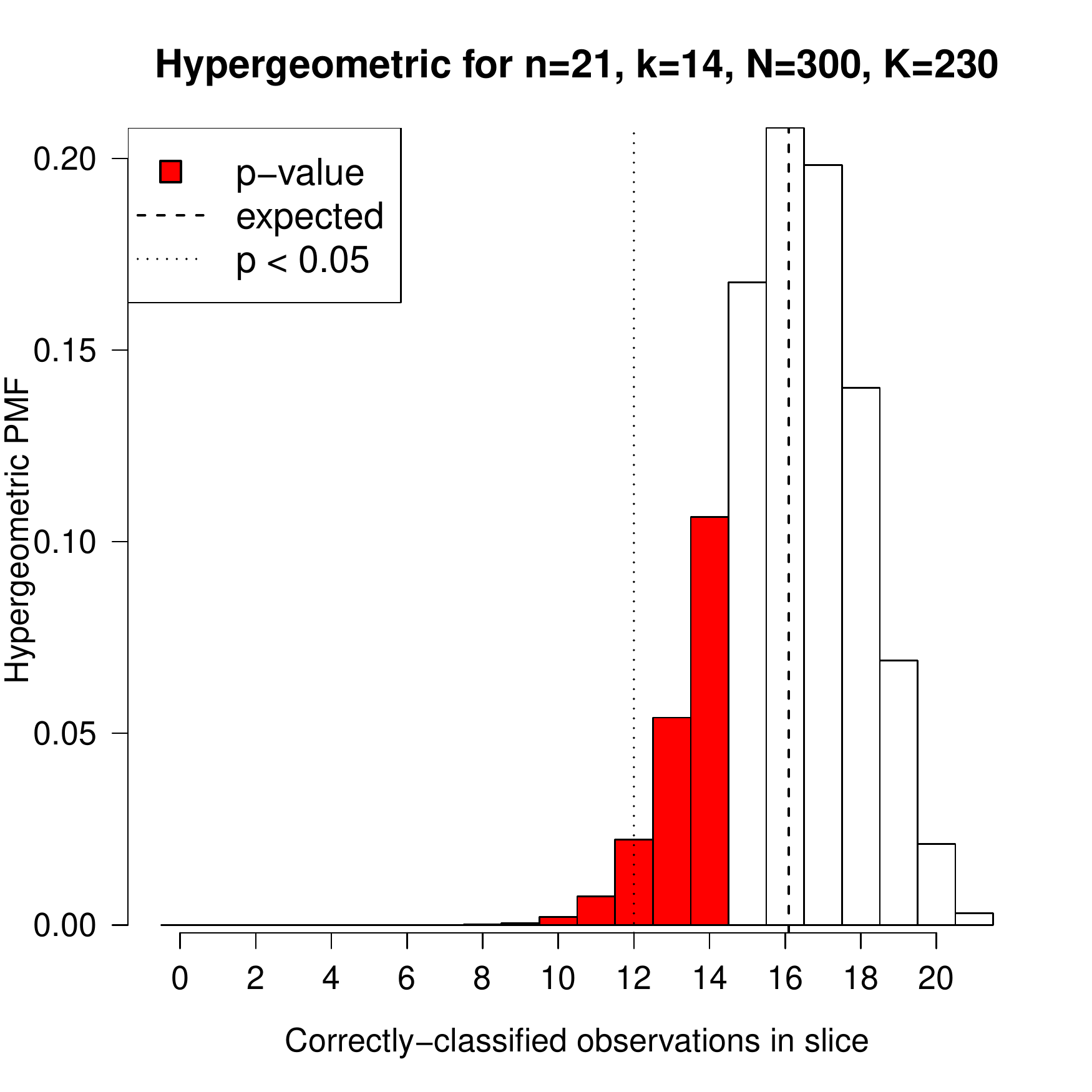}
  \caption{Hypergeometric distribution of correct classifications and p-value ($\approx 0.193$) for a slice of size $n=21$ based on the `\textit{credithistory}' feature of the \textbf{statlog} dataset.}
  \label{fig:pvalue}
\end{figure}

Figure ~\ref{fig:pvalue} illustrates the hypergeometric calculation for the \textbf{statlog} on the slice of observations with feature value `\textit{credithistory}'=5 (not shown in this paper).  The slice contains $n=21$ observations, of which $k=14$ (66.7\%) are correctly classified; the dataset overall has $N=300$ observations, of which $K=230$ (76.7\%) are correct. The hypergeometric distribution, which models the probability of $x$ correct classifications out of $n$, in this case has support on $x=\{0,\dots,21\}$. The dashed line (at 16.1, or 76.7\% of 21) indicates the number of correct classifications that would be expected for a random draw of size $n=21$, that is, the $k$ giving same accuracy as on the test set overall.  

Since we are only dealing with cases where the performance is less than expected ($k$ observed $< n\frac{K}{N}$), the p-value (shown in red in Figure ~\ref{fig:pvalue}) is the area to the left of the observed value $k=14$.  The performance is not below the expected value significantly enough, however, as the p-value $\approx 0.193$.  In order to achieve a significant p-value (below 0.05), we would need $k\leq 12$ (accuracy at most 57.1\%), indicated by the dotted line.  As discussed below, among slices with the same accuracy $k/n$, the hypergeometric p-value gives more significance to larger slices ($n$ larger).  Thus, a slice where $k=28$ and $n=42$---both doubled from before, thus leaving the slice accuracy, and the overall $K$ and $N$ unchanged---would have a p-value of $\approx 0.04$ and \textit{would} be significant.  To have the same significance, in terms of the p-value, as the actual slice, this larger slice would have about 29 correct classifications (69\%), that is, \textit{more than twice} the observed $k=14$.  This means that our original slice of size $n=21$ and accuracy 66.7\% is about as significantly as bad as a slice twice as large but with a slightly better accuracy of 69\%.

\subsubsection{Potential issues in computing significant slices}
Our slicing heuristics require the slices to consist of \textit{all} the records with the set of given categorical feature values, or of all records with numeric feature values in a given range, or to be a higher-order interaction of such single-feature slices.  By virtue of this restriction, not all possible random subsets of size $n$ of records would qualify as a slice, if there were omitted observations that would occupy the same range along one of the features.  The ML performance on an unconstrained random sample would be on average the same as that on the test set overall.  It is possible that, this constraint on the relevant subsets inherently results in ML model performances on the slice that are typically more different from the overall performance than the typical random subset is.  
This may be investigated in further research. Another issue we will investigate is how to integrate multiple hypothesis testing procedures on the slices.  This may need to account for dependence between the model's performance on difference slices due to overlap in the features included, to ensure that slices we identify as significant are so in fact.  However, we believe that slices that are filtered out as insignificant using this criterion are definitely not worth investigating further, so this hypergeometric p-value can be considered an initial sanity check on the results.

\subsubsection{Selecting slices with more support}
The procedure of filtering candidate slices based on their low p-values tends to select the larger slices among the candidates as those with worse model performance than the dataset overall.  This is a good characteristic of our algorithm, since larger slices  tend to indicate more fundamental data issues.  Small slices are more likely to represent artefacts of overfitting or simply less-interesting areas of model weakness.  

The tendency to select larger slices is a direct result of the hypergeometric distribution, specifically the fact it is based on sampling without replacement from a finite sample (in contrast to the binomial distribution).  Given a fixed dataset, the variance of the proportion of correct classifications (that is, of $X/n$ when $X$ is a hypergeometric-distributed random variable) decreases as the sample size ($n$, here the slice size) increases.  Therefore, for two slices with the same performance (i.e., proportion of  correct classifications), the larger will be a more significant result; the performance can even be slightly better in the larger slice while still being more significant than the smaller one.  For instance, in Table \ref{tab:slice_ex} on the \textbf{Adult} dataset, the slice based on `\textit{capital\_gain}' has worse model accuracy (0.688) than that based on the `\textit{age}' feature (0.797), but has a much smaller slice size ($n$=122 vs 8,538).  Nevertheless, despite the fact that accuracy on the \textit{age} slice is better, the larger size makes it a more significant instance of model under-performance, with a much lower p-value, though in this instance both slices have low p-values.  A illustrative example of this property is given in the discussion of Figure ~\ref{fig:pvalue}.

In addition to the variance aspect, which would hold in sampling with replacement, the same level of under-performance is also more significant in a larger slice when sampling without replacement because the slice size is nearer to the finite sample of correctly-classified observations.  Thus, the statistical rationale of the p-value reinforces our logical sense that larger diagnostic slices of data are more significant and helpful to the user.
\section{Discussion}\label{sec:discussion}
We report on initial promising results of automatically suggesting under-performing data slices.

In our experiments we applied FreaAI on ML test data.
Data slices in general and especially ones automatically generated such as by FreaAI are useful throughout the ML solution development life-cycle. We are working on utilizing data slices from the very early stages of ML solution development, starting from the design of a solution that provides value. We believe that our technology has other useful use cases. One use case may be to identify potential features or feature combinations for de-biasing. Bias is a prejudice that is considered to be unfair, such as refusing a loan based on the gender or race of a person. Even though FreaAI works on existing data which may already be biased and therefore the ML performance on it might be within bounds, FreaAI may occasionally identify data slices that are indicative of bias. Table~\ref{tab:frea_vs_bias} provides anecdotal evidence of that.
\begin{table}[htbp]
  \centering
  \caption{Anecdotal relation between FreaAI analysis and features identified as requiring de-biasing. The first two rows are taken from \textbf{statlog} in which age$<$25 was identified as a biasing factor and the next 3 rows from \textbf{ProPublica} where number of priors ($>1$) biases the model. The last two rows show this captured by the DT method which aggregated several values, while the other rows where detected by the HPD method.}
    \begin{tabular}{l@{\hspace{.7\tabcolsep}}r@{\hspace{.7\tabcolsep}}r@{\hspace{.7\tabcolsep}}r@{\hspace{.7\tabcolsep}}r@{\hspace{.7\tabcolsep}}}
    \toprule
    \textbf{ATTR} & \textbf{VALUE} & \textbf{SUP} & \textbf{PERF} & \textbf{p-value} \\
    \midrule
    age                 & 21                & 4     & 0.250  & 0.038 \\
    age                 & 22                & 7     & 0.428  & 0.045 \\
                        &                   &      &       &       \\
    Number\_of\_Priors  & 4                 & 119   & 0.588 & 7E-05 \\
    Number\_of\_Priors  & 5                 & 79    & 0.658 & 0.024 \\
    Number\_of\_Priors  & 3                 & 132   & 0.659 & 0.007 \\
                        &                   &      &       &       \\
    age                 & (19, 20, .., 29)   & 124   & 0.670 & 1E-04 \\
    Number\_of\_Priors  & (2, 3, 4, 5)      & 519   & 0.660 & 9E-07 \\
    \bottomrule
    \end{tabular}%
  \label{tab:frea_vs_bias}%
\end{table}%
Another use case is to improve the overall ML solution as a result of analyzing the data slices. One may add training data as characterized by data slices or add alternative logic for inputs that belong to specific slices. Computing data slices performance on the training data may assist in doing that. Table~\ref{tab:frea_on_train} provides an example. Slices that have low performance even on the training set and have low support may be indicative of a need for more training data with the slice's characteristics. The first row is an example of that. Slices that have low performance on the training data but high support, such as the second row, may be indicative of a need for alternative logic for input belonging to such slices.
\begin{table}[htbp]
  \centering
  \caption{Running FreaAI on the \textbf{training} dataset we sometimes encounter slices which under-perform (relative to the train CI) even in this dataset. The higher the model performance the fewer such slices are found. The first two rows in the table are from the \textbf{Avila} dataset and the last two from \textbf{Anuran (Family)}. The \textbf{LOW CI} column is the lower bound of the training dataset confidence interval}
    \begin{tabular}{l@{\hspace{.7\tabcolsep}}r@{\hspace{.7\tabcolsep}}r@{\hspace{.7\tabcolsep}}r@{\hspace{.7\tabcolsep}}r@{\hspace{.7\tabcolsep}}r@{\hspace{.7\tabcolsep}}r@{\hspace{.7\tabcolsep}}}
    \toprule
    \textbf{ATTR} & \textbf{VALUE} & \textbf{SUP} & \textbf{PERF} & \textbf{p-value} & \textbf{LOW CI} \\
    \midrule
    F4          & -0.633--0.634     & 12    & 0.960 & 6E-14 & 0.9992 \\
    F3          & -2.172--0.18      & 172   & 0.988 & 0     & 0.9992 \\
    MFCCs\_4    & 0.783-1           & 45    & 0.977 & 0     & 0.9986 \\
    MFCCs\_9    & -0.587--0.311     & 98    & 0.989 & 3E-05 & 0.9986 \\
    \bottomrule
    \end{tabular}%
  \label{tab:frea_on_train}%
\end{table}%

FreaAI can automatically find only issues for which data exists. We highly recommend adding manually-defined validation features to address the challenge of identifying missing requirements.

There is a challenge of combining and reporting on multiple data slices. We are experimenting with ways to optimize the different factors, such as a Pareto front, to take into account support vs performance, size vs range of a slice and more. We suspect that the use case for which the slices are being analyzed also affects the kind of slices that are more interesting. For example, if our aim is to identify features that should be de-biased, or if it is to direct the labeling effort for getting more training data, we may expect the set of interesting slices suggested by the analysis in these cases to differ from one another.

Using multiple slicing heuristics (e.g., HPD or DT) may result in overlapping slices. This creates a challenge of whether and how to combine the overlapping slices.

We experimented with extracting data slices of up to 3-way feature interactions. It is algorithmically possible to extract slices of any n-way interaction (up to the number of data features). However, we suspect this is not useful because the smaller the dataset is, the more fragmentation the interactions result in and the smaller the likelihood that data exists.
    
We experimented with multiple heuristics to suggest data slices, and more methods exist. We are exploring ways to rank the resulting slices.
Architecturally, to simplify the design, we could use the same method for identifying both 1-way and higher-order interaction slices. This could be done with the DT heuristic, for example. However, it may be the case that different slicing heuristics capture different aspects of unmet requirements. We plan to experiment with prioritizing slices generated with various techniques and develop metrics, including ones based on user feedback, to better estimate the variety of unmet requirements captured.

\section{Conclusions} \label{sec:conclusions}
We address the challenge of finding under-performing data slices to validate the ML performance of ML solutions.
We developed automated slicing heuristics and implemented them in FreaAI, such that the resulting slices are correct, statistically significant and explainable. 
We experimented with seven open datasets, demonstrating that FreaAI is able to consistently find under-performing data slices. Moreover, the reported slices often have substantial support, increasing the usefulness of the report to the user. 
\small{
\bibliography{main} 
}

\end{document}